# Adaptive boosting with dynamic weight adjustment

**Vamsi Sai Ranga Sri Harsha Mangina[1]**

[1]*Masters in Computer Science, Clemson University, Clemson, SC*

---***---

**Abstract -** *Adaptive Boosting with Dynamic Weight Adjustment is an enhancement of the traditional Adaptive boosting commonly known as AdaBoost, a powerful ensemble learning technique. Adaptive Boosting with Dynamic Weight Adjustment technique improves the efficiency and accuracy by dynamically updating the weights of the instances based on prediction error where the weights are updated in proportion to the error rather than updating weights uniformly as we do in traditional Adaboost. Adaptive Boosting with Dynamic Weight Adjustment performs better than Adaptive Boosting as it can handle more complex data relations, allowing our model to handle imbalances and noise better, leading to more accurate and balanced predictions. The proposed model provides a more flexible and effective approach for boosting, particularly in challenging classification tasks.*

## 1. INTRODUCTION

Adaptive boosting is one of the powerful ensemble learning techniques in machine learning that aims to improve the performance of weak learners by combining them into a strong classifier. The principle of Ensemble learning is to combine several weak learners into a single strong learner by using one of the Bagging, Boosting, and Stumping techniques. The AdaBoost is a boosting technique of ensemble learning. In boosting, each learner is trained with complete data rather than splitting the data in the bagging technique of ensemble learning. The principle behind the AdaBoost is to train weak learners sequentially. At first learner, each instance is assigned equal initial weights and the instances' weights are modified uniformly only once per iteration where iteration or round means a single weak classifier. Each successive learner or subsequent learner focuses on the outcome of the previous learners; It focuses mainly on the misclassified instances, thereby increasing the weights of the misclassified instances and decreasing the weights of correctly classified instances, enhancing the overall efficiency of the model.

Despite its success, the traditional AdaBoost algorithm has limitations, especially when dealing with complex datasets and finding complex relationships among the features. In such datasets, uniform changing of weights where increasing the weights of misclassified instances and decreasing the weights of correctly classified instances can be suboptimal and increase the prediction error rate of the model. There may be chances of underfitting and overfitting for AdaBoost.

To improve the efficiency and flexibility of a model handling more complex data and complex relationships among the data, we can use Adaptive Boosting with Dynamic Weight Adjustment. Adaptive Boosting with Dynamic Weight Adjustment is an enhancement of the traditional AdaBoost technique where the weight updation process in Adaptive Boosting with Dynamic Weight Adjustment is more adaptive by taking classification errors and the overall error distribution and based on the individual instances. This enables our model to work with multiclass and more complex data efficiently, enhancing the performance and its efficiency compared to the traditional AdaBoost technique. The findings suggest that dynamic weight adjustment offers a flexible and effective way to enhance boosting algorithms, making them more suitable for a broader range of classification tasks.

## 2. LITERATURE SURVEY

The most important step in the software development process is the literature review. This will describe some preliminary research that was carried out by several authors on this appropriate work and we are going to take some important articles into consideration and further extend our work.

**Freund Y., & Schapire,** R. E. (1997). "A Decision-Theoretic Generalization of On-Line Learning and an Application to Boosting. "This foundational paper introduces AdaBoost, an adaptive boosting algorithm where misclassified examples are given higher weights in subsequent rounds. The algorithm adjusts weights dynamically to focus on harder cases. AdaBoost was shown to improve the performance of weak learners significantly.

**Rätsch G., Onoda T., & Müller,** K.-R. (2001). "Soft Margins for AdaBoost." Machine Learning, 42(3), 287-320. This paper extends AdaBoost by introducing the concept of soft margins, allowing for better handling of noisy data. The dynamic adjustment of weights is refined to balance the trade-off between model complexity and margin errors.

**Dietterich,** T. G. (2000). "An Experimental Comparison of Three Methods for Constructing Ensembles of Decision Trees: Bagging, Boosting, and Randomization." Machine Learning, 40(2), 139-157. This experimental study compares boosting with other ensemble methods and highlights the efficacy of dynamic weight adjustment in boosting. The paper provides empirical evidence on how weight adjustment in boosting leads to improved classification performance.

**Zhu J., Zou H., Rosset S., & Hastie,** T. (2009).

"Multi-class AdaBoost." Statistics and Its Interface, 2(3), 349-360. The authors extend AdaBoost to multi-class problems and introduce techniques for dynamic weight adjustment tailored for multi-class classification. The paper discusses the theoretical underpinnings and provides empirical results showing the effectiveness of these adjustments.

**Wang J., & Sun,** Y. (2013).

"Boosting for learning multiple classes with imbalanced class distribution." 2013 IEEE 13th International Conference on Data Mining. This work addresses the challenge of class imbalance in boosting algorithms. It proposes dynamic weight adjustment strategies to give more focus to minority classes, improving the overall classification performance in imbalanced datasets.

**Friedman,** J. H. (2001).

"Greedy Function Approximation: A Gradient Boosting Machine." Annals of Statistics, 29(5), 1189-1232. Although primarily about gradient boosting, this paper lays the groundwork for understanding how boosting methods, including those with dynamic weight adjustment, can be formulated as a gradient descent optimization problem. It provides insights into the theoretical aspects of weight adjustment in boosting.

**Bauer E., & Kohavi,** R. (1999).

"An Empirical Comparison of Voting Classification Algorithms: Bagging, Boosting, and Variants." Machine Learning, 36(1-2), 105-139. This empirical study compares various boosting algorithms and their variants, focusing on how dynamic weight adjustment impacts performance. The findings emphasize the importance of weight adjustment in achieving superior results.

**Ridgeway,** G. (1999).

"The State of Boosting." Computing Science and Statistics, 31, 172-181. A review of the state of boosting techniques at the time, this paper discusses various adaptations of boosting algorithms, including those with dynamic weight adjustment. It provides a comprehensive overview of the strengths and limitations of different boosting methods.

## 3. Existing System :

### AdaBoost :

The AdaBoost (Adaptive Boosting) algorithm, introduced by Freund and Schapire in 1997, is a widely used ensemble learning method that combines multiple weak classifiers to create a strong classifier. Here's a detailed overview of the existing system:

**Initialization :**

Assign equal weights to all training samples:

$W_i = 1/N$, where N is the number of samples.

**Training Process :**

**Iteration**: Repeat for T=1,2....,T(where T is the number of weak classifiers):

**Train Weak Classifier**: Train a weak classifier $h_t$ using the weighted training samples.

**Calculate Error**: Compute the weighted error rate $\epsilon_t$ of the weak classifier:

$$\epsilon_t = \frac{\sum_{i=1}^{N} w_i \mathbb{I}(h_t(x_i) \neq y_i)}{\sum_{i=1}^{N} w_i}$$

Where $\mathbb{I}$ is the indicator function.

**Classifier Weight:** Calculate the weight $\alpha_t$ of the classifier:

$$\alpha_t = \frac{1}{2} \ln\left(\frac{1 - \epsilon_t}{\epsilon_t}\right)$$

**Update Sample Weights**: Adjust the weights of the training samples:

$$w_i \leftarrow w_i \cdot \exp(\alpha_t \cdot \mathbb{I}(h_t(x_i) \neq y_i))$$

### Combination of Weak Classifiers

The final string classifier is a weighted majority vote of the weak classifiers:

$$H(x) = \text{sign}\left(\sum_{t=1}^{T} \alpha_t h_t(x)\right)$$

Here, \sign denotes the sign function which returns the class label.

**The weights are updated as shown in the figure :**

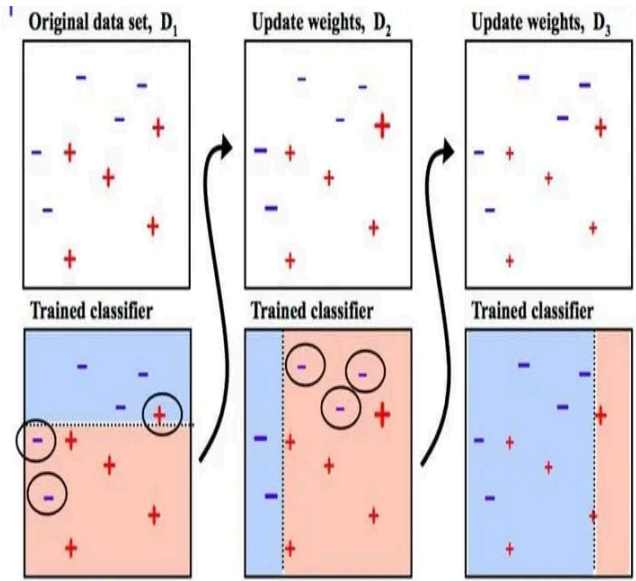

Figure 3.1: Weight Updation process in AdaBoost

## 4. Proposed System :

The proposed system aims to improve the standard AdaBoost algorithm by incorporating dynamic weight adjustment mechanisms, gradient boosting techniques, and enhanced handling of multi-class and imbalanced datasets. Here's a detailed elaboration:

**Adaptive Weight Adjustment :**

In the standard AdaBoost algorithm, weights of incorrectly classified samples are increased exponentially, while weights of correctly classified samples are decreased. The proposed system introduces more sophisticated strategies for adjusting these weights dynamically:

- **Soft Margins:** Following the work of Ratsch, Onoda, and Müller (2001), the proposed system can use soft margin techniques. This approach adjusts weights based on a margin that allows some errors but penalizes them less severely, thus making the classifier more robust to noisy data.

- **Confidence-Rated Predictions:** As Schapire and Singer (1999) suggested, incorporating confidence-rated predictions means that the classifier outputs not just the class label but also a confidence level. This confidence level can then be used to adjust the weights more accurately.

For each round t:

- **Initial Weights**: $w_i^{(t)}$ are initialized equally.

- **Update Weights :**

$$w_i^{(t+1)} = w_i^{(t)} \cdot \exp(\alpha_t \cdot \mathbb{I}(h_t(x_i) \neq y_i))$$

where $\alpha_t$ is the confidence of the weak classifier $h_t$, and $\mathbb{I}$ is the indicator function.

**Confidence-Rated Predictions :**

**Classifier Confidence :**

$$\alpha_t = \frac{1}{2} \ln \left( \frac{1 - \epsilon_t}{\epsilon_t} \right)$$

where $\epsilon_t$ is the weighted error rate of the weak classifier $h_t$

**Gradient Boosting Integration :**

Gradient boosting, introduced by Friedman (2001), focuses on optimizing a differentiable loss function. The proposed system integrates gradient boosting concepts to refine the weight adjustment process:

**Loss Function Optimization:** Instead of simply minimizing classification error, the system can minimize a more sophisticated loss function, such as logistic or exponential loss, leading to better generalization and performance.

$$L(y, F(x)) = \sum_{i=1}^{N} \ell(y_i, F(x_i))$$

where $\ell$ is a differentiable loss function (e.g., logistic loss).

**Updation Rule :**

$$F_{m+1}(x) = F_m(x) + \gamma_m h_m(x)$$

where $\gamma_m$ is the step size, and $h_m$ is the weak learner.

**Boosting Framework:** Each weak learner is trained to correct the errors of the previous learners in a gradient descent manner, making the overall model more accurate and robust.

# 5. Result Analysis :

**Utilizing Google Colab for Model Performance Evaluation**

Google Colab is a free platform for Python and machine learning tasks, offering limited but sufficient resources. Here, we analyze our models' performance using Google Colab. The steps involved are as follows :

## 5.1 Open Google Colab :

**Connect to a GPU or CPU Environment :**
Start by selecting the appropriate runtime (GPU or CPU) depending on the computational requirements of your task.

**Upload the Necessary Datasets :**
Load the datasets into your Colab notebook from your local system or directly from online repositories.

## 5.2 Ensure to import essential libraries such as :

import numpy as np
import pandas as pd
import sklearn
from sklearn import preprocessing
from sklearn.model_selection import train_test_split
from sklearn import metrics
from sklearn.ensemble import AdaBoostClassifier
from sklearn.naive_bayes import GaussianNB

## 5.3 Data Preprocessing :

**Replace Missing Values**: Handle any missing data to maintain data integrity.
**Remove Unnecessary Features**: Drop features that do not contribute to the model's performance.
**Eliminate Duplicate Values**: Ensure there are no redundant data entries.
**Transform Categorical Feature**: Convert categorical data into numerical formats.
**Note:** This is a crucial step, as the quality of data preprocessing significantly affects model performance.

## 5.4 Split the Dataset :

Divide the dataset into training and testing sets.
**Training Data:** Used to train the model.
**Testing Data:** Used to evaluate the model's performance.

## 5.5 Compare Model Performance :

We compared the performance of two models :
- AdaBoost
- AdaBoost with Dynamic Weight Adjustment

## 5.6 Model Analysis :

For both AdaBoost and AdaBoost with Dynamic Weight Adjustment, GaussianNB (a probabilistic classifier and a variant of Naive Bayes) was used as the base estimator. We analyzed the models' performance using the following datasets :
- Rice Variants
- Dry Bean

## 5.7 Performance Metrics and Observations :

**AdaBoost :**

**Rice Variants :**

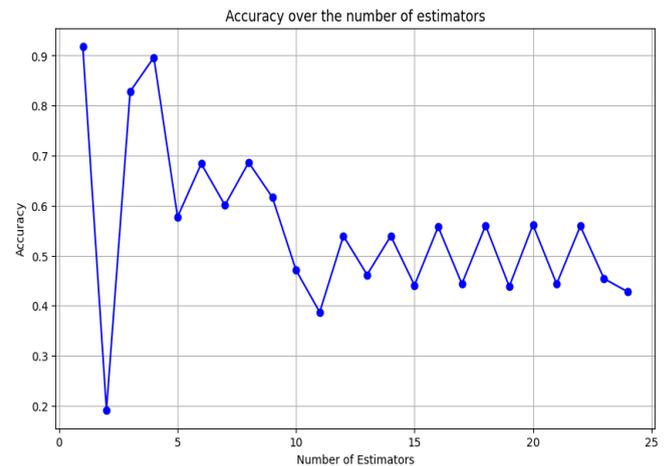

Figure 5.7.1: Accuracy Graph for Rice Variants Over Adaboost

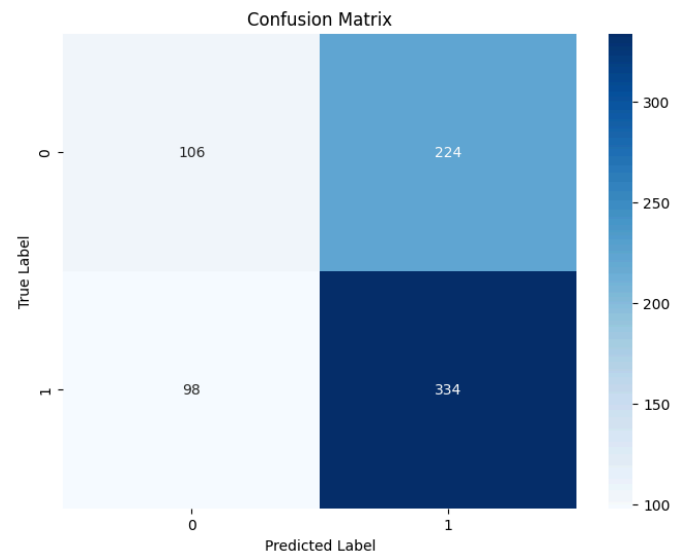

Figure 5.7.2: Confusion Matrix for Rice Variants Over Adaboost

Overall Accuracy(Rice variants) : 0.5774278215223098

- **Dry Bean:**

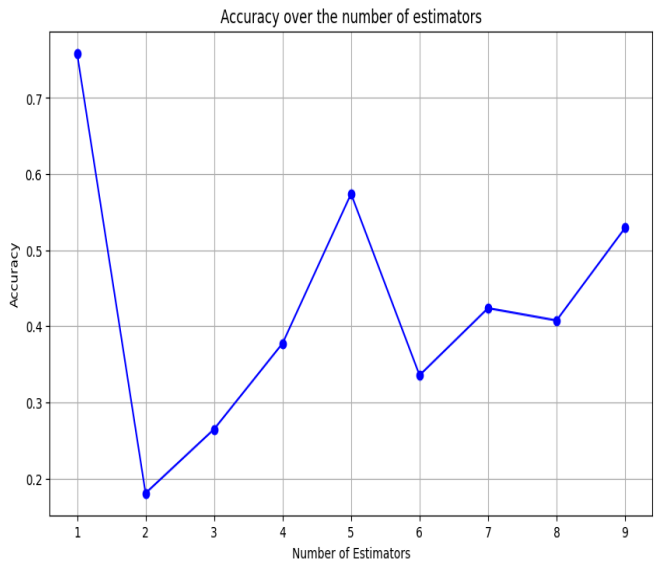

Figure 5.7.3: Accuracy Graph for Dry Bean Over Adaboost

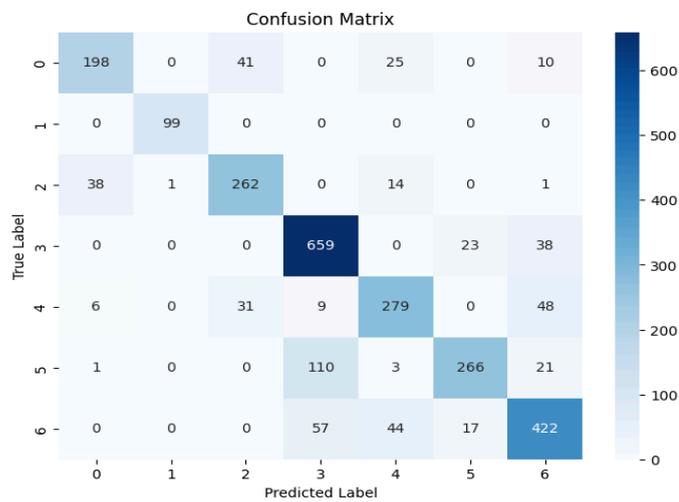

Figure 5.7.4: Confusion Matrix for Dry Bean Over Adaboost

Overall Accuracy(Dry Bean) : 0.18068307014322438

**AdaBoost with Dynamic Weight Adjustment**

- **Rice Variants :**

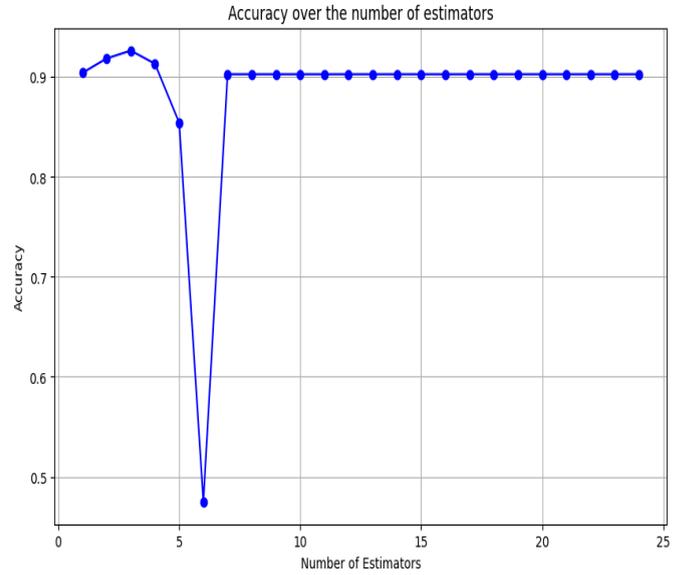

Figure 5.7.5: Accuracy Graph for Rice Variants Over Adaboost with Dynamic Weight Adjustment

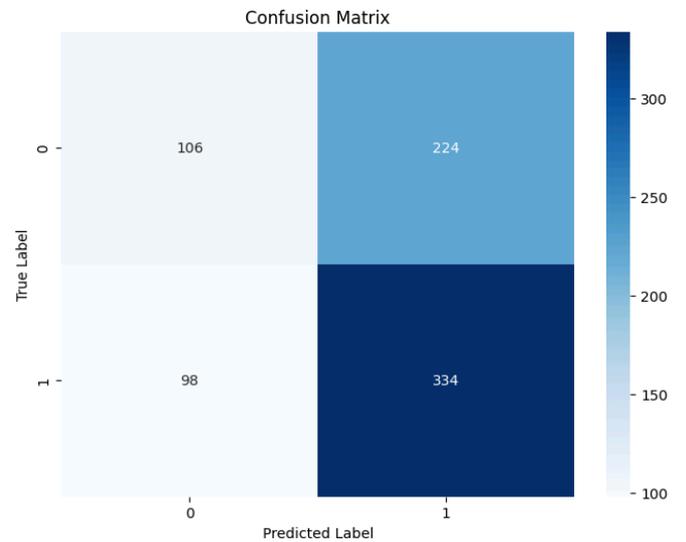

Figure 5.7.6: Confusion Matrix for Rice Variants Over Adaboost with Dynamic Weight Adjustment

Overall Accuracy(Rice variants) : 0.8571428571428571

● **Dry Bean:**

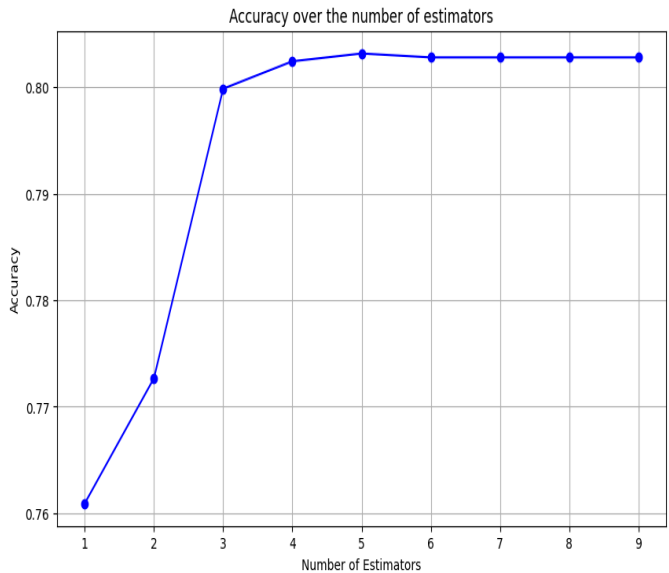

Figure 5.7.7: Accuracy Graph for Dry Bean Over Adaboost with Dynamic Weight Adjustment

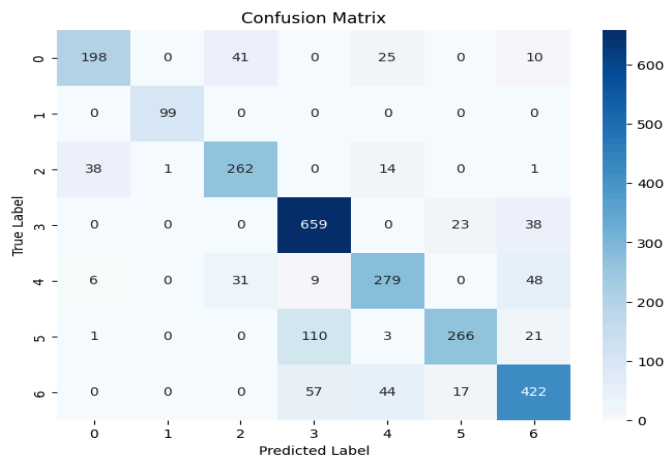

Figure 5.7.8: Confusion Matrix for Dry Bean Over Adaboost with Dynamic Weight Adjustment

Overall Accuracy(Dry Bean) : 0.8024237972824091

**Traditional AdaBoost :**

The learning curve indicates significant fluctuations in accuracy over iterations, demonstrating instability. This instability suggests that traditional AdaBoost struggles to maintain consistent performance with varying numbers of weak learners.

The model required a higher number of weak learners to achieve stable performance.

**Adaptive Boosting with Dynamic Weight Adjustment :**

The learning curve shows a rapid increase in accuracy, reaching a stable performance with fewer weak learners. This indicates that the dynamic weight adjustment mechanism helps the model to focus more effectively on misclassified instances, leading to faster convergence and stable performance.

The model achieved stability with fewer weak learners, highlighting the efficiency of the dynamic weight adjustment approach.

## 6. CONCLUSIONS

The proposed Adaptive Boosting with Dynamic Weight Adjustment demonstrates superior performance over traditional AdaBoost, particularly in accuracy and efficiency. The ability to achieve stable and high performance with fewer estimators makes it a robust and effective alternative for binary and multi-class classification tasks. This enhanced boosting technique is particularly beneficial for complex datasets with diverse class distributions and intricate feature relationships. The dynamic weight adjustment mechanism ensures that the model can adapt more effectively to difficult instances, leading to better generalization and overall performance.